# Machine Vision System for Early-stage Apple Flowers and Flower Clusters Detection for Precision Thinning and Pollination

**Salik Ram Khanal\*, Ranjan Sapkota, Dawood Ahmed, Uddhav Bhattarai, Manoj Karkee\***

*\* Corresponding Author (e-mail: salik.khanal@wsu.edu, manoj.karkee@wsu.edu)*
*Center for Precision and Automated Agricultural Systems, Biological Systems Engineering, Washington State University, Prosser WA USA*

**Abstract:** Early-stage identification of fruit flowers that are in both opened and unopened condition in an orchard environment is significant information to perform crop load management operations such as flower thinning and pollination using automated and robotic platforms. These operations are important in tree-fruit agriculture to enhance fruit quality, manage crop load, and enhance the overall profit. The recent development in agricultural automation suggests that this can be done using robotics which includes machine vision technology. In this article, we proposed a vision system that detects early-stage flowers in an unstructured orchard environment using YOLOv5 object detection algorithm. For the robotics implementation, the position of a cluster of the flower blossom is important to navigate the robot and the end effector. The centroid of individual flowers (both open and unopen) was identified and associated with flower clusters via K-means clustering. The accuracy of the opened and unopened flower detection is achieved up to mAP of 81.9% in commercial orchard images.

*Keywords:* Agriculture automation, Precision Thinning, Flower Detection, Flower Clustering,

## 1. INTRODUCTION

The United States (U.S.) specialty crop production contributes to 30% to 40% of the total U.S. crop value (USDA, 2019). However, it faces various challenges such as labor shortages for manual flower thinning, and bee shortages for pollination, resulting in a big threat to global food security. Since agricultural robots have the potential to replace/reduce human labor, there is a dire need for the development of automated and robotic platforms for crop load management in agriculture (Bochtis et al., 2020).

Crop load management in tree fruit production is a balancing act of reducing crop load in the current season for desired fruit size and quality and achieving adequate return bloom for the coming season (Terence Robinson and Hoying, 2016). The machine-vision-based robotic crop load management system could have the potential to perform automated flower thinning, pollination, and green fruit thinning in a real-time and natural environment. Among these, pollination and flower removal activities can be performed at the same time. In addition, the same information opens up the door for the development of a robotic flower removal system (Ren and Yang, 2016; Iwanami et al., 2018). The planning strategies of crop load based on early-stage flowers could be one of the most effective ways to manage crop load in modern orchards. Therefore, a robust system for the identification of early-stage flowers could be a significant pathway to automated crop load management.

Most of the recent articles proposed algorithms that could only be used for flowers that resemble the full bloom conditions of flowers. But there are very few or no concrete studies that have been reported regarding the identification of flower blossoms during the early growing season. As precise crop load management operation starts from the early season; it is crucial to have information related to flower clusters of opened and unopened flowers. Additionally, the identification of a king flower in a cluster during the early season could be significant information to perform precise and automated pollination. Many researchers are reported that the You Only Look Once (YOLO) object detection algorithm performs better accuracy and faster speed in custom object detection as compared with Region-Based Convolutional Neural Network (RCNN) based object detection. Therefore, in this paper, the YOLO object detection technique (YOLOV5) is used to detect both open and unopened flowers. However, to our knowledge, there is not any proposed robotic vision system that could perform automated flower thinning and precise pollination using a robotic platform during the early growing season of apples. To overcome these limitations, the main objectives of this study are:

- to detect unopened and open flower clusters during the early flowering season using deep learning techniques.
- to associate individual flower detection to flower cluster using the k-means algorithm.

## 2. RELATED WORKS

Various machine vision technologies have been applied using various types of object detection algorithms in agriculture automation for more than a decade. Nilsback and Zisserman (2006) proposed a visual vocabulary that can support object classification for flowers that have a significant visual similarity using traditional image processing methods. Traditional image processing techniques consist of various operations such as image enhancement, image restoration, and image analysis that can be performed to manipulate images using digital computers. Another early study on feature extraction of lesquerella flower was presented by Thorp and Dierig (2011), where the authors developed an image processing algorithm to detect flowers using image segmentation by transforming the image to the hue, saturation, and intensity (HSI) color space. Hočevar et al. (2014) estimated the number of flower clusters of individual trees in a high-density apple orchard by implementing apple flower detection based on thresholding and morphological image processing in hue, saturation, luminance (HSL) color space image. Likewise, there are several studies that performed flower detection using the traditional image processing methods such as morphological image processing and various segmentation techniques including Otsu's method, and classifying the flower color group with contour shapes using k-means clustering (Biradar and Shrikhande, 2015; Hong and Choi, 2012; Tiay et al., 2014).

However, the accuracy of this traditional method is relatively low compared to deep learning algorithms, and the approach is limited to specific scenarios such as the requirement of enough daylight and using artificial background such as a black cloth screen behind the trees to adjust illuminance making the system only applicable in a controlled environment. Additionally, these methods could not take morphological features into account which caused the requirement of adjusting thresholding parameters whenever changes in illumination, flowering density, and camera position occur. On the other hand, these techniques have their applicability impeded especially by variable lighting conditions and occlusion by leaves, flowers, or stems (Gongal et al., 2015). Moreover, most of these traditional techniques could not identify situations like the difference between buds and flowers, and flower overlapping for accurate production estimation.

To overcome these challenges, Machine Learning (ML), for the last few years has become one effective way to process a large amount of data in agriculture. Tran et al. (2018) developed a flower and visitor (bees, insects, etc.) detection system in unconstrained conditions, where the author made use of Convolutional Neural Networks (CNN) for both flower-based image segmentation and visitor detection. However, the method generated a higher flower misdetection rate (8.12%) which further resulted in a higher visitor misdetection rate. Safar and Safar (2019) proposed another intelligent flower detection system based on an ML model "ResNet", using model enhancements such as fine-tuning, dropout ratio, and class weight to modify and improve the accuracy of the ResNet for flower detection. Likewise, Islam et al. (2020) proposed a CNNbased flower detection system for eight varieties of flowers using activation functions "ReLu" and "softmax", and optimizer function "Adam". More studies such as (de Luna et al., 2020; Yahata et al., 2017; Zawbaa et al., 2014) presented flower detection in tomato, soybean, and eight flower dataset respectively using ML classification techniques such as Simple Linear Iterative Clustering (SLIC), Support Vector Machine (SVM) and Random Forest (RF) in last few years. However, the accuracy reported from these studies is relatively lower.

Most of the research studies in agriculture automation are specific to the plant or the product. Limited studies are proposed in state-of-art focused on the apple. Dias et al. (2018) recently developed a DL-based approach for apple flower detection using CNN which was pre-trained for saliency detection. The existing network has been finetuned by combining CNN and SVM together to become flower sensitive. Cheng and Zhang (2020) have proposed a flower detection system for smart gardens based on a deep learning model called You Only Look Once (YOLOv4), where the author applied CSPDarnet53 network as a backbone network to reduce network computation and increase the speed of flower detection. Patel (2020) proposed flower classification approach using an optimized deep CNN by integrating Neural Architecture Search-Feature Pyramid Network (NAS-FPN) with Faster Region-based Convolutional Neural Network (Faster R-CNN), and using transfer learning based on COCO dataset (Li et al., 2022). Using the

YOLOv4 object detection model and fine-tuning, various flower detection models are proposed (Zhou et al., 2022; Matsui et al., 2009). Bhattarai and Karkee (2022) recently developed a regression-based neural network, also a weakly supervised approach called CountNet, that detects and counts the apple flower to estimate bloom density, crop load, and yield. However, all these recent studies for flower detection using DL are only validated for the flowers that exhibit full bloom conditions.

## 3. MATERIALS AND METHODS

### 3.1 Data Acquisition and Data Preparation

The 2D RGB images for this study were collected in a commercial apple orchard in Prosser, Washington, USA using Intel Realsense 435i camera (Santa Clara, California, USA). The images were collected during the early bloom season from April 9 – April 20, 2022. The unopened and king bloom condition of the flower clusters was visually assessed for six different sessions and image images were collected during each session. The images were collected in the early bloom season and objects were classified into two classes - opened and unopened flowers. The image dataset contains 529 images with more than 5000 labels (around 90% unopened and 10% opened) of classes. The images were labeled with LabelImg annotation application.

### 3.2 Proposed Approach

Object detection is one of the key outcomes of deep learning algorithms. Many object detection algorithms have been proposed and come into use in the last decade and each algorithm has specific features. The YOLO (You Only Look Once) object detection algorithm family is one of the popular algorithms because of its processing speed. In this study, the YOLOv5 object detection algorithm was used to detect early-stage apple flowers which include both unopened and opened flowers as the different categories, and the results are post-processed to find the flower clusters. Based on the centroid of the flower clusters, the robotic system can be navigated to perform desired field operations.

The proposed system consists of two steps: early-stage flower detection and clustering. In machine learning algorithms, the parameters used in machine learning models are optimized during model training and validation. In the YOLOv5 algorithm (Jocher et al., 2022), the hyperparameters can be optimized using parameter optimization (Hutter et al., 2019). The 25 hyperparameters were optimized in ten iterations during the training phase. For further experiments, the optimized values of the hyperparameters were used.

After detecting the unopen and open flowers, the centroid of each detected bounding box was calculated, followed by the k-means algorithm without defining the value of $k$ as the number of clusters in each image frame can vary. $k$ value varies from 1 to $n$, where $n$ is the maximum number of clusters in an image frame. The best value of the $k$ was calculated using Silhouette analysis (Rousseeuw, 1987). The Silhouette coefficient $S(i)$ is calculated using this equation.

$$S(i) = \frac{b(i) - a(i)}{max(a(i), b(i))} \quad (1)$$

where $S(i)$ is the silhouette coefficient of the data point $i$, $a(i)$ is the average distance between $i$ and all the other data points in the cluster to which $i$ belong, and $b(i)$ is the average distance from $i$ to all clusters to which $i$ does not belong.

After detecting each cluster and its centroid, each cluster is assigned a cluster identity, so that the decision-making system can decide on the next cluster after completing the thinning or pollination in each cluster. The overall proposed methodology is illustrated in Figure 1.

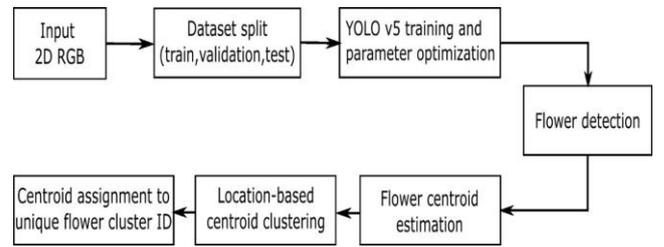

Fig. 1. Block diagram of the proposed algorithm

### 3.3 Data Preparation and Model Training

Based on the number of convolutional layers in the architecture, YOLOv5 has five different models represented by N, S, M, L, and X which stand for Nano, Small, Medium, Large, and extra-large, respectively. The performance of the models is compared to our dataset. The important parameters to evaluate the models are execution speed, accuracy, mAP, etc. For further experiments, the best models were chosen based on both accuracy and execution speed for further data analysis.

All the images are separated into training, validation, and testing in the ratio of 80:10:10. Using the training dataset, all the experiments were carried out in 300 epochs with an image size of 640x640 and the batch size of 16. The algorithm is evaluated using recall, precision, mAP, etc.

$$recall = \frac{True\ Positive}{True\ Positive + False\ Negative} \quad (2)$$

$$precision = \frac{True\ Positive}{True\ Positive + False\ Positive} \quad (3)$$

$$mAP = \frac{1}{N}\sum_{i=1}^{N} AP_i \quad (4)$$

### 3.4 Clustering Technique and Implementation Plan

After detecting the early-stage flowers, the next step is to find flower clusters so that we can find the centroid point of each cluster to locate the end-effector of the thinning or pollination robot. The overall algorithm for the clustering is given below.

---
**Algorithm 1** Proposed Algorithm

**Require:** Flower image frame **for**
  each frame **do**
    Detect flowers and calculate their centroids
    Apply k-means clustering technique: find the flower clusters. Initialize: number of max items in a cluster = m. maximum number of clusters = n. **for** n= 1 to n **do**
      Calculate Silhouette coefficient (s).
      k = k[i] where max[s[i]]].
    Apply k-means clusters with k.
    Capture next frame **end**
  **for**

---

### 4. EXPERIMENTS AND RESULTS

The first step was to find the best model among five different YOLOv5 models. The selection was based on both accuracy and execution speed. In our experiments, the heaviest model (YOLOv5X) was too large and does not fit in our GPU, so it is excluded. The rest of the four models were trained. The results of the experiments with four different models of YOLOv5 are illustrated in Table 1. Based on the mAP@0.5, the best result is obtained from the YOLOv5s model with a reasonable execution speed. For further experiments, YOLOv5s will be used.

Figure 2 shows the result of object detection. The early stage flowers are represented by an orange bounding box and the open flower is represented by a dark red bounding box. The model is also able to detect overlapping flowers. From the results, the detection accuracy is promising for robotics operations such as flower thinning and pollination.

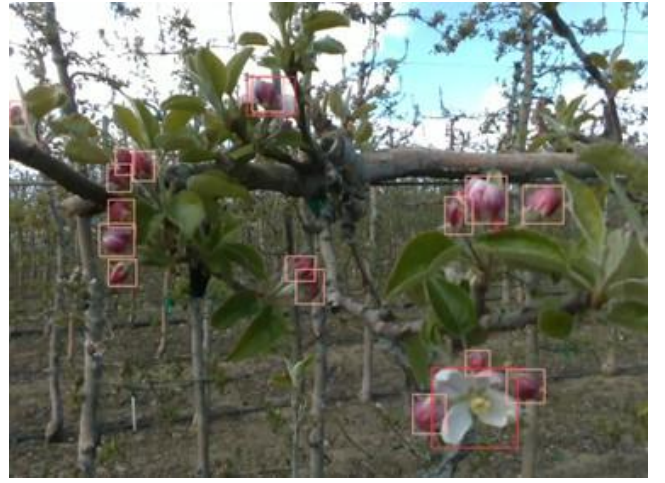

Fig. 2. Illustration of the open and unopened flower detection using YOLOv5 object detection model.

The most important and final step of the experiments is to find the clusters of flowers and find the centroid of each cluster. Figure 3 shows the results of the association of individual flower detections to unique flower clusters and estimated flower cluster centroid. The centroid of each cluster is an important reference point set up the end-effector of the pollination or thinning robot.

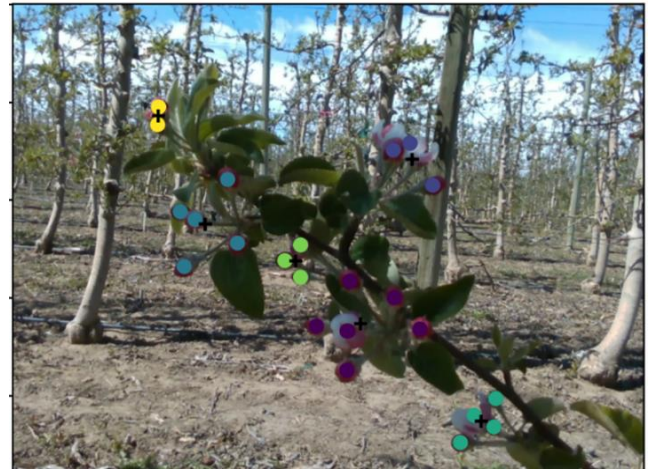

Fig. 3. Detection and clustering of the unopened flowers. Each (+) represents the centroid of each cluster.

### 5. DISCUSSION

One of the initial steps in the apple orchard is apple flower thinning for quality fruit production using proper crop load management. In recent years, much research has proposed flower-thinning robots and the most important part of this type of robot is machine vision. In this article, the early-stage flowers where most flowers were not opened were detected to conduct efficient flower thinning and pollination.

Table 1. Matrices of the different models of the YOLOv5 object detection models.

| Models | mAP@0.5 | Precision | Recall |
|---|---|---|---|
| YOLOv5n | 0.72 | 0.65 | 0.82 |
| **YOLOv5s** | **0.819** | **0.735** | 0.835 |
| YOLOv5m | 0.762 | 0.68 | 0.881 |
| YOLOv5l | 0.767 | 0.709 | 0.79 |

Overall processing for early-stage flower detection is more accurate than the open flowers. Using the machine vision system and flower cluster geometry, the king flower can also be detected which might be the best choice to select the position or orientation of the end effector. Unopen flowers are separated from each other and are easy to detect. In agriculture robotics automation, the motion and all the movements are controlled by the output of the vision system. Most of the research articles on flower thinning and pollination consider only open flower detection, which is not fair as not all the flowers open at the same time. This article covers a wide range of open and unopen flower detection.

In this article, the centroid of each cluster of flowers is calculated so that we can position the end-effectors of the robot to conduct thinning and pollination. Especially, the design of the robot end-effector is easier with the unopened flower, and easier to remove unwanted flower buds. The detection evaluation matrices indicate the results are quite enough to implement in the field. The mAP of the bud is 0.819. In the case of clustering, the accuracy is quite enough, and we don't need to define the k-value in k-means.

The important contribution is the proposed algorithm for the implementation where the proposed algorithm is not limited to flower thinning. Some protocols or algorithms can be applied to pollination, green apple thinning, and harvesting. According to [20], the control of the robot's pose, and position is based on the detected object and repositioned for the next operation.

This research study has a few limitations and future work. The following are the limitations of this result study. 1) All the experiments were carried out using RGB images, so for the real implementation, we need to consider the 3D information. The next proposed step is to work on pose estimation. 2) All the experiments were carried out using flower images collected in the same orchard, the results might be different in different orchards and image data collection timing. To design a universal model, diverse types of images are necessary to collect.

## 6. CONCLUSIONS

In this article, both open and unopen flowers are detected using the YOLOv5 algorithm and applied k-clustering algorithm to find the centroid of each cluster. The centroid location of each cluster is important to locate the end-effector of the robot. The efficacy of the vision system in thinning and pollination robots is enough to implement in the field environment. In automation robotics in agriculture, the motion and movement of the robot are controlled according to the 3D position of the object and robot. The future direction of this study is to work with 3D pose estimation of the flowers or blossoms.

## ACKNOWLEDGEMENTS

This work was funded by Agricultural AI for Transforming Workforce and Decision Support (AgAID).

## REFERENCES

Bhattarai, U. and Karkee, M. (2022). A weakly-supervised approach for flower/fruit counting in apple orchards. *Computers in Industry*, 138, 103635.

Biradar, B.V. and Shrikhande, S.P. (2015). Flower detection and counting using morphological and segmentation technique. *Int. J. Comput. Sci. Inform. Technol*, 6, 2498–2501.

Bochtis, D., Benos, L., Lampridi, M., Marinoudi, V., Pearson, S., and Sørensen, C.G. (2020). Agricultural workforce crisis in light of the covid-19 pandemic. *Sustainability*, 12(19). doi:10.3390/su12198212. URL https://www.mdpi.com/2071-1050/12/19/8212.

Cheng, Z. and Zhang, F. (2020). Flower end-to-end detection based on yolov4 using a mobile device. *Wireless Communications and Mobile Computing*, 2020.

de Luna, R.G., Dadios, E.P., Bandala, A.A., and Vicerra, R.R.P. (2020). Tomato growth stage monitoring for smart farm using deep transfer learning with machine learning-based maturity grading. *AGRIVITA, Journal of Agricultural Science*, 42(1), 24–36.

Dias, P.A., Tabb, A., and Medeiros, H. (2018). Apple flower detection using deep convolutional networks. *Computers in Industry*, 99, 17–28.

Gongal, A., Amatya, S., Karkee, M., Zhang, Q., and Lewis, K. (2015). Sensors and systems for fruit detection and localization: A review. *Computers and Electronics in Agriculture*, 116, 8–19.

Hoˇcevar, M., Sirok, B., Godeˇsa, T., and Stopar, M. (2014). Flowering estimation in apple orchards by image analysis. *Precision Agriculture*, 15(4), 466–478.

Hong, S.W. and Choi, L. (2012). Automatic recognition of flowers through color and edge based contour detection. In *2012 3rd International conference on image processing theory, tools and applications (IPTA)*, 141–146. IEEE.

Hutter, F., Kotthoff, L., and Vanschoren, J. (2019). *Automated Machine Learning - Methods, Systems, Challenges*. Springer Nature. doi:10.1007/978-3-030-05318-5.

Islam, S., Foysal, M.F.A., and Jahan, N. (2020). A computer vision approach to classify local flower using convolutional neural network. In *2020 4th International Conference on Intelligent Computing and Control Systems (ICICCS)*, 1200–1204. IEEE.

Iwanami, H., Moriya-Tanaka, Y., Honda, C., Hanada, T., and Wada, M. (2018). A model for representing the relationships among crop load, timing of thinning, flower bud formation, and fruit weight in apples. *Scientia Horticulturae*, 242, 181–187.

Jocher, G., Chaurasia, A., Stoken, A., Borovec, J., NanoCode012, Kwon, Y., TaoXie, Fang, J., imyhxy, Michael, K., Lorna, V, A., Montes, D., Nadar, J., Laughing, tkianai, yxNONG, Skalski, P., Wang, Z., Hogan, A., Fati, C., Mammana, L., AlexWang1900, Patel, D., Yiwei, D., You, F., Hajek, J., Diaconu, L., and Minh, M.T. (2022). ultralytics/yolov5: v6.1 - TensorRT, TensorFlow Edge TPU and OpenVINO Export and Inference. doi:10.5281/zenodo.6222936. URL https://doi.org/10.5281/zenodo.6222936.

Li, G., Suo, R., Zhao, G., Gao, C., Fu, L., Shi, F., Dhupia, J., Li, R., and Cui, Y. (2022). Real-time detection of kiwifruit flower and bud simultaneously in orchard using yolov4 for robotic pollination. *Computers and Electronics in Agriculture*, 193, 106641.

Matsui, T., Suzuki, S., Ujikawa, K., Usui, T., Gotoh, S., Sugamata, M., Badarch, Z., and Abe, S. (2009). Development of a non-contact screening system for rapid medical inspection at a quarantine depot using a laser doppler blood-flow meter, microwave radar and infrared thermography. *Journal of medical engineering & technology*, 33(5), 403–409.

Nilsback, M.E. and Zisserman, A. (2006). A visual vocabulary for flower classification. In *2006 IEEE Computer Society Conference on Computer Vision and Pattern Recognition (CVPR'06)*, volume 2, 1447–1454. doi:10.1109/CVPR.2006.42.

Patel, Isha, S. (2020). An optimized deep learning model for flower classification using nas-fpn and faster r-cnn. *International Journal of Scientific & Technology Research*, 9(03), 5308–5318.

Ren, D. and Yang, S. (2016). Intelligent automation with applications to agriculture. *Intelligent Automation & Soft Computing*, 22, 227–228. doi:10.1080/10798587.2015.1095473.

Rousseeuw, P.J. (1987). Silhouettes: A graphical aid to the interpretation and validation of cluster analysis. *Journal of Computational and Applied Mathematics*, 20, 53–65. doi:https://doi.org/10.1016/0377-0427(87)90125-7.

Safar, A. and Safar, M. (2019). Intelligent flower detection system using machine learning. In *Proceedings of SAI Intelligent Systems Conference*, 463–472. Springer. Terence Robinson, Alan Lakso, D.G. and Hoying, S. (2016). Precision Crop Load Management. https://nyshs.org/wp-content/uploads/2016/10/Pages-6-10-from-NYFQ-Summer-Book-6-22-2013. PRESS-2.pdf.

Thorp, K. and Dierig, D. (2011). Color image segmentation approach to monitor flowering in lesquerella. *Industrial crops and products*, 34(1), 1150–1159.

Tiay, T., Benyaphaichit, P., and Riyamongkol, P. (2014). Flower recognition system based on image processing. In *2014 Third ICT International Student Project Conference (ICT-ISPC)*, 99–102. IEEE.

Tran, D.T., Høye, T.T., Gabbouj, M., and Iosifidis, A. (2018). Automatic flower and visitor detection system. In *2018 26th European Signal Processing Conference (Eusipco)*, 405–409. IEEE.

USDA (2019). Specialty crops. https://www.nass.usda.gov/Publications/AgCensus/2017/Online_Resources/Specialty_Crops/SCROPS.pdf.

Yahata, S., Onishi, T., Yamaguchi, K., Ozawa, S., Kitazono, J., Ohkawa, T., Yoshida, T., Murakami, N., and Tsuji, H. (2017). A hybrid machine learning approach to automatic plant phenotyping for smart agriculture. In *2017 International Joint Conference on Neural Networks (IJCNN)*, 1787–1793. IEEE.

Zawbaa, H.M., Abbass, M., Basha, S.H., Hazman, M., and Hassenian, A.E. (2014). An automatic flower classification approach using machine learning algorithms. In *2014 International conference on advances in computing, communications and informatics (ICACCI)*, 895–901. IEEE.

Zhou, H., Wang, X., Au, W., Kang, H., and Chen, C. (2022). Intelligent robots for fruit harvesting: Recent developments and future challenges. *Precision Agriculture*, 1–52.